\documentclass[letterpaper]{article} 
\usepackage{aaai24}  
\usepackage{times}  
\usepackage{helvet}  
\usepackage{courier}  
\usepackage[hyphens]{url}  
\usepackage{graphicx} 
\usepackage{natbib}  
\usepackage{caption} 
\usepackage{algorithm}

\usepackage{textcomp}
\usepackage{xcolor}
\usepackage{mdframed}
\usepackage{comment}
\usepackage{tikz}
\usepackage{pifont}
\usepackage{algpseudocode}
\newcommand{\vect}[1]{\boldsymbol{#1}}

\usepackage{newfloat}
\usepackage{colortbl}
\newcommand{\myTitle}{LeanContext\;}
\usepackage{listings}
\DeclareCaptionStyle{ruled}{labelfont=normalfont,labelsep=colon,strut=off} 
\floatstyle{ruled}
\newfloat{listing}{tb}{lst}{}
\floatname{listing}{Listing}
\usepackage{amssymb,amstext,amsmath,newtxmath}
\usepackage{booktabs}
\nocopyright 

\title{LeanContext: Cost-Efficient Domain-Specific Question Answering Using LLMs}
\author{
    Md Adnan Arefeen\textsuperscript{\rm 1,2} \footnote{Work done during internship at NEC Laboratories America}, 
    Biplob Debnath\textsuperscript{\rm 1}, and 
    Srimat Chakradhar\textsuperscript{\rm 1}
}
\affiliations{
    \textsuperscript{\rm 1}NEC Laboratories America, \textsuperscript{\rm 2} University of Missouri-Kansas City\\
        aa4cy@umsystem.edu, \{biplob, chak\}@nec-labs.com
}

\usepackage{bibentry}

\begin{document}

\maketitle

\begin{abstract}
Question-answering (QA) is a significant application of Large Language Models (LLMs), shaping chatbot capabilities across healthcare, education, and customer service. However, widespread LLM integration presents a challenge for small businesses due to the high expenses of LLM API usage. Costs rise rapidly when domain-specific data (context) is used alongside queries for accurate domain-specific LLM responses. One option is to summarize the context by using LLMs and reduce the context. However, this can also filter out useful information that is necessary to answer some domain-specific queries. In this paper, we shift from human-oriented summarizers to AI model-friendly summaries. Our approach, \myTitle, efficiently extracts \emph{k} key sentences from the context that are closely aligned with the query. The choice of \emph{k} is neither static nor random; we introduce a reinforcement learning technique that dynamically determines \emph{k} based on the query and context. The rest of the less important sentences are reduced using a free open source text reduction method. We evaluate \myTitle against several recent query-aware and query-unaware context reduction approaches on prominent datasets (arxiv papers and BBC news articles). Despite cost reductions of $37.29\%$ to $67.81\%$, \myTitle's ROUGE-1 score decreases only by $1.41\%$ to $2.65\%$ compared to a baseline that retains the entire context (no summarization). Additionally, if free pretrained LLM-based summarizers are used to reduce context (into human consumable summaries), \myTitle can further modify the reduced context to enhance the accuracy (ROUGE-1 score) by $13.22\%$ to $24.61\%$.

\end{abstract}

\section{Introduction}

In recent times, large language models (LLMs) have seen extensive utilization, especially since the introduction of LLM APIs for customer-oriented applications on a large scale~\cite{liu2023summary}. These applications include chatbots (like GPT-4), language translation~\cite{jiao2023chatgpt}, text summarization~\cite{luo2023chatgpt,yang2023exploring,zhang2023extractive}, and question-answering (QA) tasks~\cite{tan2023evaluation}, personalized robot assistance~\cite{wu2023tidybot}. While the zero-shot performance of the LLM model is nearly on par with fine-tuned models for specific tasks, it has limitations. One significant limitation is its inability to answer queries about recent events on which it has not been trained. This lack of exposure to up-to-date information can lead to inaccurate responses, particularly for domain-specific information processing, where the LLMs may not grasp new terminology or jargon. To build an effective domain-specific question-answering system, it becomes essential to educate the LLMs about the specific domains, enabling them to adapt and understand new information accurately.

\begin{figure}[!tpb]
    \centering
    \includegraphics[width=\linewidth]{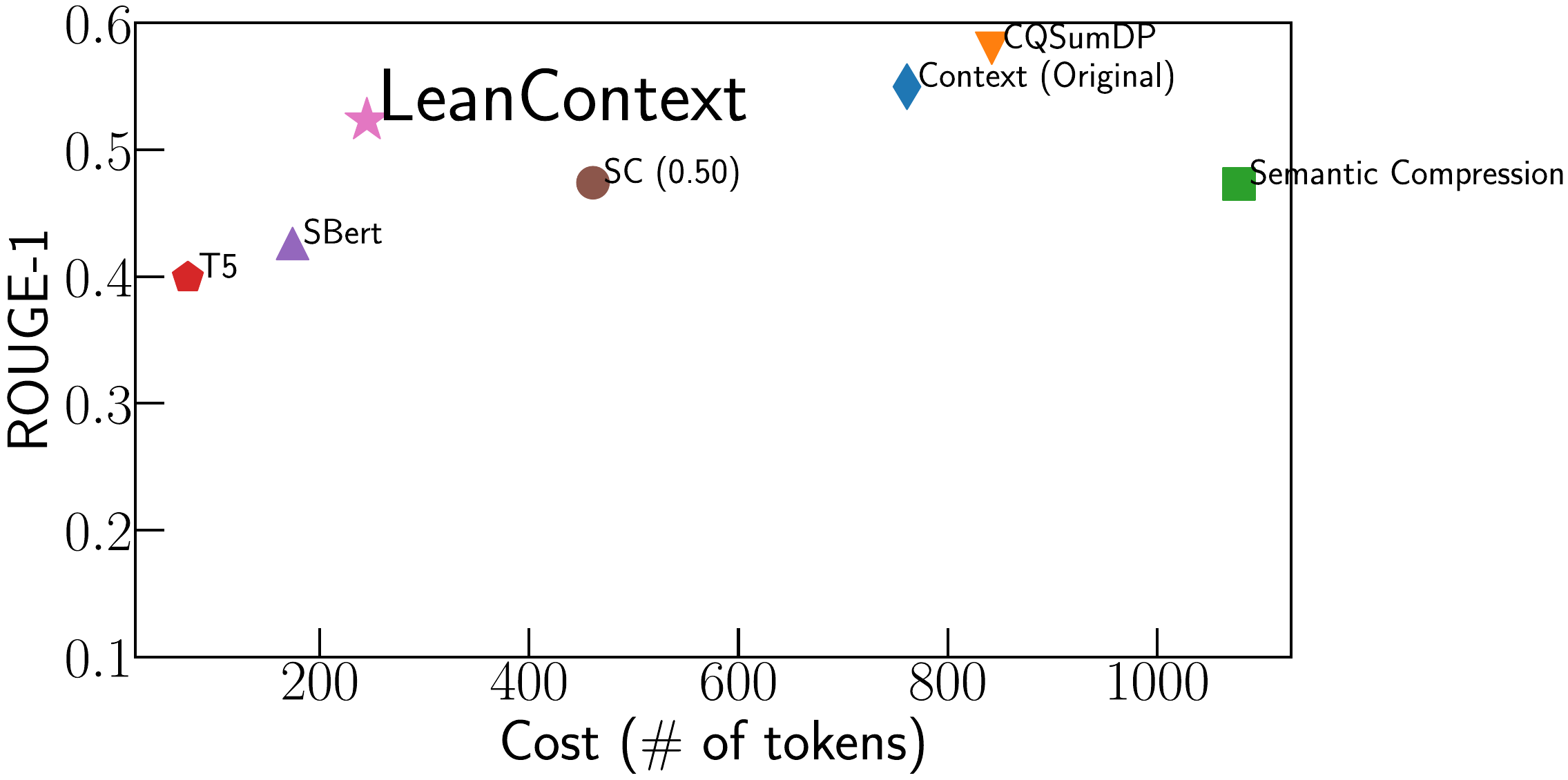}
    \caption{Compared to the original context LeanContext only drops in $\sim$2\% ROUGE-1 score with $\sim$ 68\% savings on BBCNews dataset~\cite{li2023unlocking}.}
    \label{fig:overall-cost}
\end{figure}

LLMs can learn domain-specific information in two ways, (a) via \emph{fine-tuning} the model weights for the specific domain, (b) via \emph{prompting} means users' can share the contents with the LLMs as input context. Fine-tuning these large models containing billions of parameters is expensive and considered impractical if there is a rapid change of context over time~\cite{schlag2023large} e.g. a domain-specific QA system where the documents shared by users are very recent and from different domains. A more practical way is to select the latter approach i.e. the prompt-based solution, where relevant contents from user documents are added to the query to answer based on the context. Motivated by this, we focus on prompt-based solutions for document-based QA systems. 

One of the challenges for long document processing using a prompt-based solution is the input prompt length being limited to a maximum length defined by the LLM API. The token limit of GPT-3.5 and GPT-4 vary from 4,096 to 32,768 max tokens limit proportional to the usage cost. Therefore, LLMs will fail to answer the query if the prompt length is larger than the max token limit due to the larger context length in the prompt. One suitable way to get rid of this problem is via document \emph{chunking}~\cite{langchain}. In this case, initially, the user documents are segmented into chunks. Only the relevant chunks of fixed size are retrieved as context based on the query. 


The cost for context-based querying using LLM APIs via prompting is associated with a cost that is proportional to the number of input tokens (contributing prompt cost), and the number of output tokens (contributing generation cost). According to a recent study\cite{costgpt}, with 15,000 visitors having 24 requests per month, the cost of using GPT-3 (Davinci model) is \$14,400 per month (assuming prompt tokens $=1800$, output tokens $=80$) which is challenging for a small business to operate. For GPT-4, the cost is even higher than this amount. In this paper, our focus is to reduce this cost.

To mitigate the cost of using LLM API, the number of tokens in the context should be reduced as the cost is proportional to the length of the context. A low-cost option to reduce the context is to summarize the context using free open-source summarizer models. However, for domain-specific QA applications, the pre-trained open-source summarizers do not contribute to good accuracy. On the contrary, using a pay-per-use model like ChatGPT further increases the query processing cost instead of reducing it as the additional cost is added at the time of text reduction.

To this end, we propose a domain-specific query-answering system \myTitle, where users ask queries based on a document. To answer a query, \myTitle first forms a context from the document based on the query by retrieving relevant chunks. Next, it identifies the top-$k$ sentences related to the query from the context. \myTitle  introduces a reinforcement learning technique that dynamically determines k based on the query and context. Then, \myTitle reduces the rest of the sentences in fragments by an open source text reduction method. Next, it forms a new context by stitching top-k sentences and reduced fragments in the order of their appearance order in the original context. Finally, it invokes an LLM (like ChatGPT) to answer the query using that new context. It is to be noted that the goal of \myTitle is contrary to the summarization task that generates a meaningful summary for human users.  Rather, in \myTitle, the reduced context will be consumed by a question-answering model like ChatGPT. Figure~\ref{fig:overall-cost} shows the scenario of \myTitle, reducing the context size with accuracy close to the original context and outperforming other open-source models.

In summary, \myTitle makes the following contributions:
    \begin{itemize}
    \item It presents a low-cost domain-specific QA system, which reduces the LLM API usage cost by reducing the domain context through the selection of important sentences related to the query and keeping them intact, while reducing rest of the sentences in between important sentences through a free open-source summarizer. It proposes a reinforcement learning technique to select the percentage of the important sentences.
   
    \item It reduces the LLM API usage cost by $37.29\%\sim67.81\%$ of a domain-specific QA system with little drop in performance by only $1.41\%\sim2.65\%$. (Table~\ref{tab:arxiv-results}, Table~\ref{tab:bbcnews-eval}). 
    
    
     \item It boosts the QA performance by $13.22\%\sim24.59\%$ (Table~\ref{tab:top-kadvg}) by combining query-aware top-$k$ sentences with the reduced context generated through free open-source text summarizers .

    \end{itemize}

\section{Related Work}
For domain-specific tasks, LLMs can be utilized to adapt the domains without modifying their inner parameters via discrete prompting where distinct instructions with contexts can be delivered as an input to generate responses for downstream tasks~\cite{ling2023beyond,brown2020language}. For domain-specific QA tasks, the domain can be reduced by context summarization to reduce the LLM cost. A lot of research has been conducted for summarizing text~\cite{miller2019leveraging,yang2023exploring}. Existing research works can be categorized into two main parts: (a) extractive and (b) abstractive. The extractive summarizers~\cite{miller2019leveraging} first identify important sentences from the text, and next summarizes them. While abstractive summarizers~\cite{laskar2023cqsumdp} reduce the context by generating new sentences. The main goal of both approaches is to generate a meaningful summary for human users.  In contrast, the goal of \myTitle is to reduce context which will be consumed by a question-answering model like ChatGPT. For the prompt-based summarization task, recently, iterative text summarization~\cite{zhang2023summit} has been proposed to refine the summary task in a feedback-based iterative manner.
In aspect or query-based summarization~\cite{yang2023exploring} summaries are generated based on a domain set of specific queries.

Query-unaware text compression via prompting is also observed in recent literature. Semantic compression~\cite{gilbert2023semantic} involves generating systematic prompts to reduce context using ChatGPT model (GPT-3.5-turbo, GPT-4) and acquire reasonable compression compared to the zlib compression method. Due to limited context window size, recent literature focus on prompt context filtering. In selective context~\cite{li2023unlocking}, token, phrase, or sentence-level query-unaware content filtering is proposed using the entropy of GPT-2 model logits for each entity. Usage of ChatGPT model in the medical domain especially in radiology is explored via prompt-engineering~\cite{ma2023impressiongpt} to summarize difficult radiology reports. Extract-then-generate pipeline-based summarization improves abstractive summary faithfulness~\cite{zhang2023extractive} through the chain of thought (CoT)~\cite{wei2022chain} reasoning. To reduce the cost of the use of LLM, FrugalGPT~\cite{chen2023frugalgpt} proposed several ideas regarding prompt adaptation by query concatenation, LLM approximation by fine-tuning or caching, and LLM cascade by the selective selection of LLMs from lower to higher cost based on a query. Still, it lacks context compression ideas to reduce the prompt tokens.

It is to be noted that recent studies either focus on summarization as a downstream task or utilize the summary of the context for the question-answering task. For most of the existing content filtering approaches, the main focus is to query-agnostic filter content with the deletion of less informative content or for solely summarization tasks. Using LLM for query-aware context reduction adds extra overhead for using pay-per-use LLM to answer correctly. In addition, in recent articles, the chunk-based preprocessing of the article is ignored by assuming each content in the dataset as a chunk. In \myTitle, the main focus is to reduce the LLM cost by considering query-aware context reduction. Due to the possibility of rapid change of domain-specific user data, fine-tuning LLM or parameter-efficient LLM is not a feasible solution. Utilizing open-source LLM~\cite{touvron2023llama,chung2022scaling} model either does not perform well on domain data or adds additional deployment cost. Consider a small business to run, we consider using pay-per-use LLMs such as OpenAI LLMs to make the system running at a reasonable cost by reducing the context. 

\section{Domain-specific QA System}
\begin{figure}[!htbp]
    \centering
    \includegraphics[width=\columnwidth]{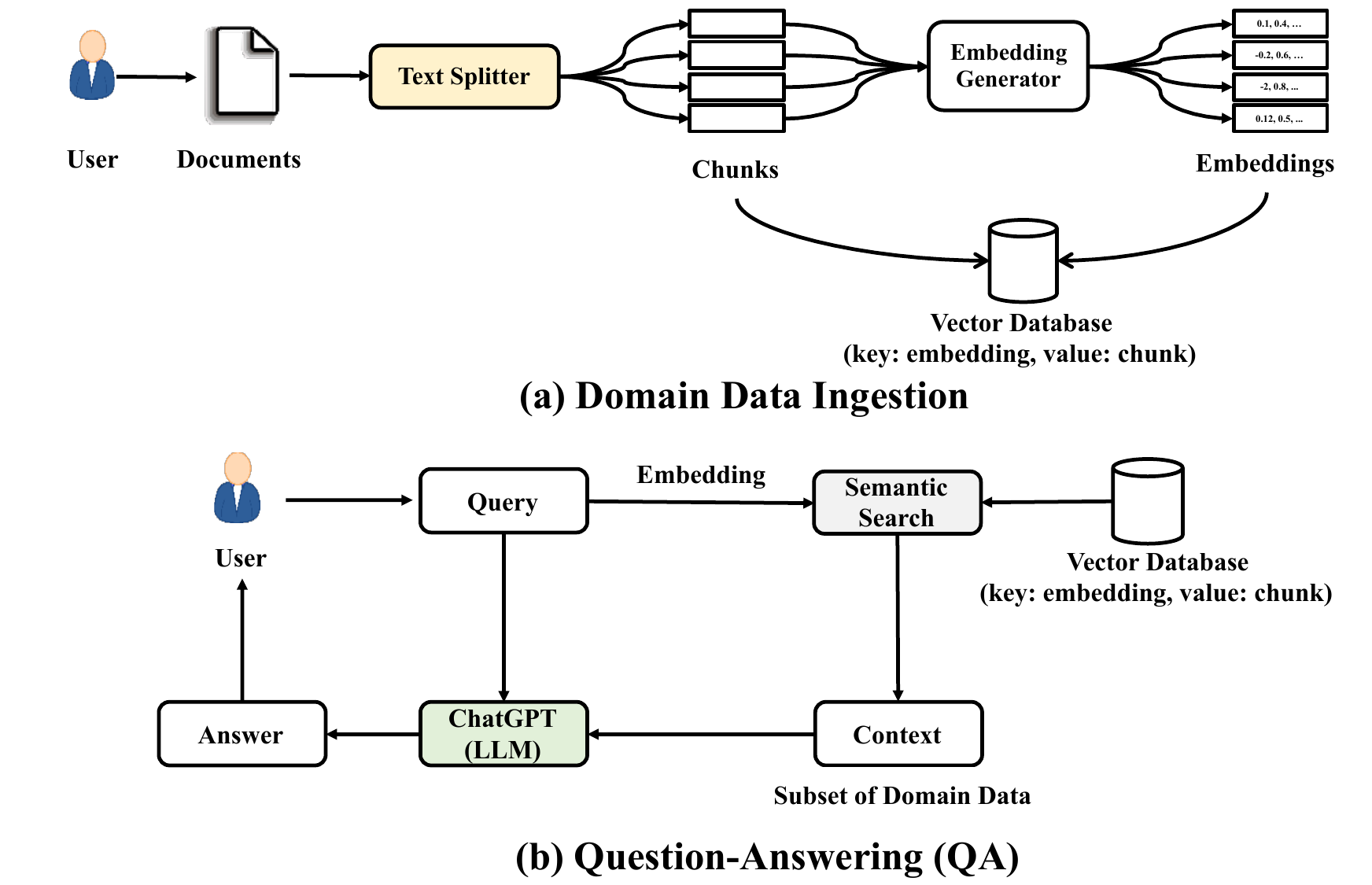}
    \caption{Workflow of a domain-specific QA system.}
    \label{fig:ingestion-qa}
\end{figure}

In a domain-specific QA system, a context with a query is given to an LLM to get the answer. If the context size exceeds the max-token limits of the LLM API, the LLM will fail to answer the query. As a result, for long document processing, a vector database~\cite{chromadb,vectordb} is used to store domain-specific documents into a number of small chunks so that a subset of relevant domain context can be retrieved from the long document rather than the whole document as context. A domain-specific QA system is shown in Figure~\ref{fig:ingestion-qa}. The QA system can be divided into two steps. 

\paragraph{(a) Domain data ingestion:} In this step, the documents, $\mathcal{D}$ will be split into a number of fixed chunks ($c$) by a text splitter. An embedding function computes the embeddings of each chunk using an embedding generator. The chunks along with the embedding vector ($\mathbf{v}_c$) of each chunk are stored in a vector database. 

\paragraph{(b) QA:} At the question-answering (QA) step, given a user query $q_i$, similar $N$ chunks are retrieved by their embeddings similar to query embedding using \emph{semantic search}. These chunks form the context ($\mathbf{C}$). Finally, the context which is a subset of domain data is fed into LLM to get the answer.

As domain-specific data and user queries are dynamic in nature, retrieving minimal context based on a query is challenging. One possible way is to make the chunk size and number of chunks dynamic so that the context contains minimal sentences to answer the query. But this solution is infeasible as the vector database needs to be reconfigured again per query with the change of domain. Instead, it will be more practical if after getting the possible chunks as context, the context is further reduced based on a query to get the near-optimal cost of LLM. Following this notion, we propose \myTitle, an adaptive context reduction system to reduce the prompt cost of ChatGPT like LLMs.

\section{\myTitle}

The objective of \myTitle is to further reduce the context $\mathbf{C}$  to $\mathbf{C'}$ where the token count of $\mathbf{C'}$ is smaller than the token count of $\mathbf{C}$. As LLM prompt cost is proportional to the token count of context, \myTitle helps to reduce the prompt cost of LLMs.
In other words, if the total number of tokens in $\mathbf{C}$ is $T$ and the number of tokens in $\mathbf{C'}$ is $t$ then \myTitle reduces the token ratio $\tau$ is defined as, $\tau = \frac{t}{T}$ without compromising the accuracy ($acc$) of the QA system. So, if the optimal accuracy of the system is $acc^*$, we formulate the optimization problem of \myTitle as follows. 
\begin{equation}
\begin{alignedat}{2}
& \text{min.} & \quad & (1-\alpha) \times \tau + \alpha \times |acc-acc^*| \\
\end{alignedat}
\end{equation}
Finally, the reduced context $\mathbf{C'}$ and the query $(q_i)$ are given to the pay-per-use LLM API to answer the query. Then, the answer is shown to the respective user via an interactive interface.

\begin{figure}[!htbp]
    \centering
    \includegraphics[width=\linewidth]{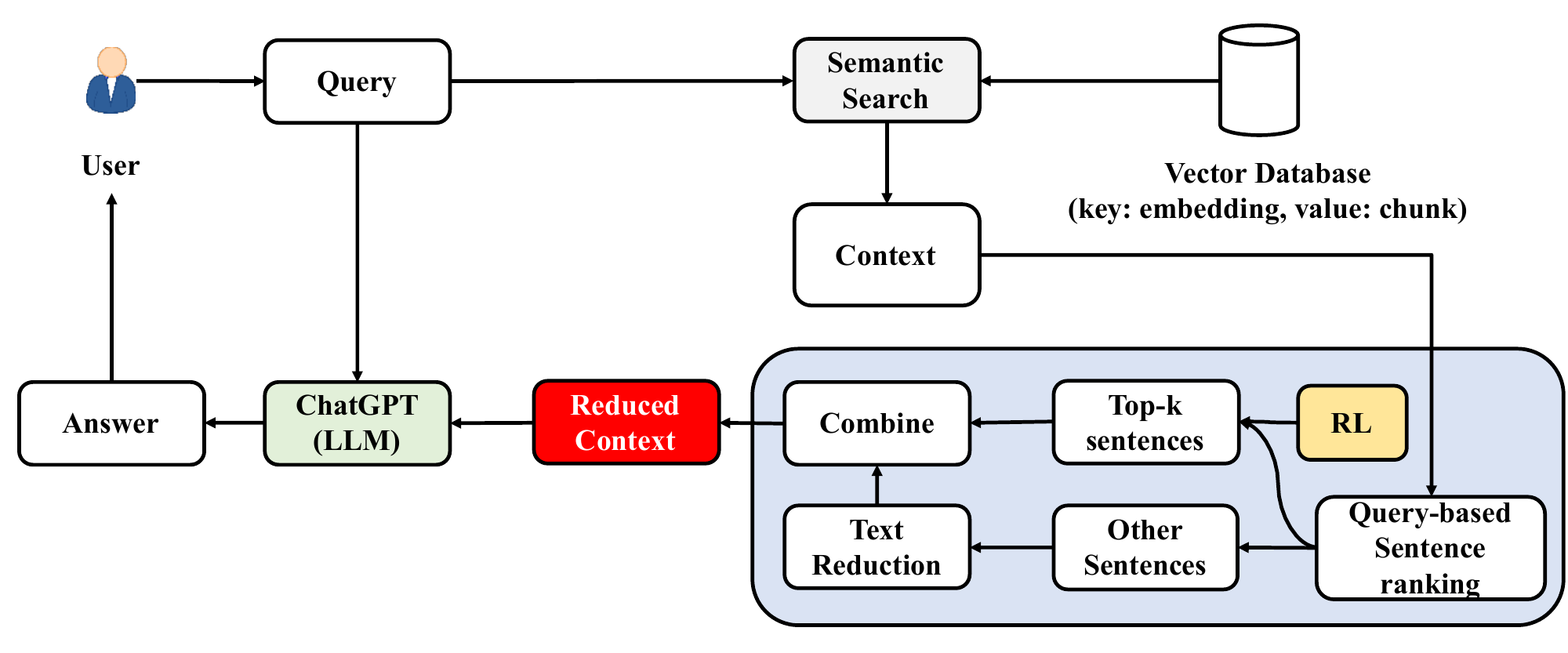}
    \caption{LeanContext System}
    \label{fig:leancontext}
\end{figure}

For context-based QA, generally, the answers reside within a couple of sentences. If the smallest amount of context for a certain question can be identified, the same response can be provided by LLMs at a lower cost i.e. less prompt tokens. So, identifying the top-$k$ sentences can reduce the context without compromising accuracy. Motivated by this simple idea, we propose \myTitle which is shown in Figure~\ref{fig:leancontext}.

\begin{figure}[!htpb]
    \centering
    \includegraphics[width=0.8\linewidth]{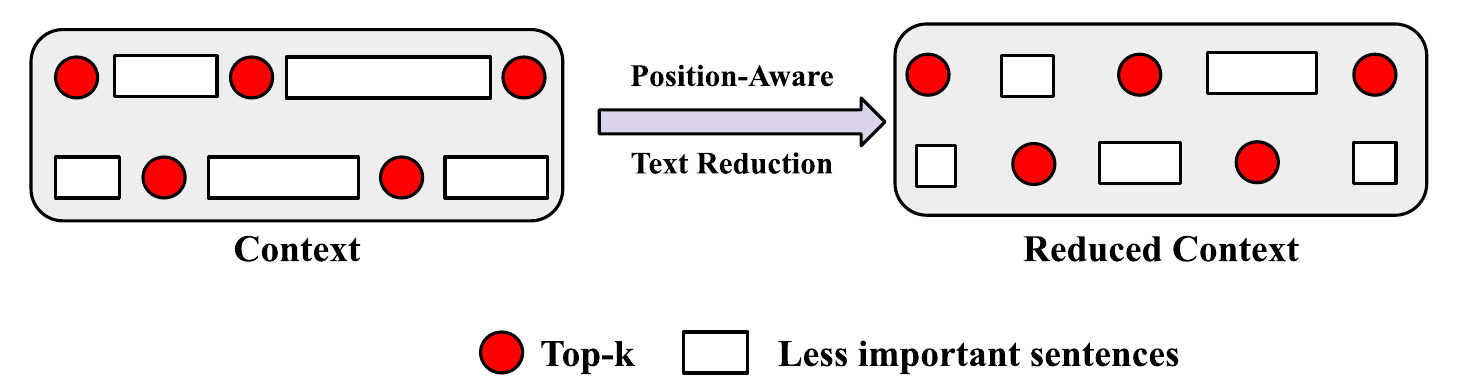}
    \caption{Illustration of the reduction of less important sentences while keeping the top-k sentences intact in a context.}
    \label{fig:order}
\end{figure}

After forming the context with semantic search, \myTitle first \emph{ranks} the sentences of the context based on a query. Assuming the context consists of a sequence of sentences ($s_1, s_2, s_3,..., s_n$), it extracts top-$k$ sentences similar to the query from context using the cosine-similarity function. To accomplish this, it computes the embedding of the query ($\mathbf{v}_q$), and the embedding is compared with each of the sentence embedding ($\mathbf{v}_{s_i}$) and \emph{top-$k$ sentences} are identified. 
\[
\text{Top-}k \;\text{sentences} = \textit{sort}(\mathbf{V}, \text{similarity\_score}(\mathbf{v}_q, \mathbf{v}_{s_i}))\\
\]
Here, $1\le i\le n$. Using only top-$k$ sentences as a reduced context is a lightweight approach to reduce the cost LLM API usages. However, \myTitle boosts accuracy by combining information from the rest of the sentences. Figure~\ref{fig:order} shows an illustration of this combination process. \myTitle keeps the top-$k$ sentences intact, while other sentences between the top-$k$ sentences are reduced by an open-source text reduction method. \myTitle maintains the order of the top-k sentences and other sentences according to their appearances in the original context in order to produce more accurate results.



The idea of maintaining the top sentences intact is a simple but interesting approach. Adding this simple approach to the existing open-source pre-trained summarizer models will help increase the performance. However, an interesting question lies in identifying the \emph{k} in top-$k$. 
We ask the question here, \emph{(a) What is the number of $k$?} 
We answer the question in a more detailed manner with a reinforcement learning (RL) based solution as follows.
\begin{algorithm}[H]
  \caption{\myTitle Training Algorithm}
  \label{algo-ls}
  \begin{algorithmic}[1]
    \Require{$\mathcal{D}$: Input documents}
    \Require{$\vect{q}$: A set of queries}
    \Require{$\Theta$: a set of predefined thresholds, acts as a set of actions in RL}
    \Ensure{$Q^*$: Trained $Q$-table}
    
    \State $\mathcal{X} \gets \emptyset$ \Comment{set of states}
    \For{each $q,\textit{context} \in \vect{q},
    \mathcal{D}$}
        \State$\mathbf{v}_{c},\mathbf{v}_{q} \gets \text{Embedding}
            (\textit{context}),\text{Embedding}(q)$  \Comment{get embedding vectors}
        \State $\mathcal{X} \gets \mathcal{X} \cup  (\mathbf{v}_{c} - \mathbf{v}_{q})$
    \EndFor
    \State $\mathcal{S} \gets \textit{K-Means}(\mathcal{X})$ \Comment{centroids as state vectors}
    \For{each $q,\textit{context} \in \vect{q},
    \mathcal{D}$}
        \State $\textit{state}\gets \text{get\_state}(\mathcal{S},\textit{context},q)$
        \State $action \gets \text{get\_action} (\Theta)$ 
            \State $\mathbf{C} \gets \text{retrieve}(q, \mathbf{v}_{q},\theta)$ \Comment context formation
            \State $\mathbf{C}' \gets \text{perform\_action}(\mathbf{C},\textit{action})$ \Comment{text reduction}
            \State $\textit{answer} \gets \textit{llm}(q,\mathbf{C}')$
            \State $\textit{r} \gets \text{compute\_score}(\textit{answer},\textit{original\_answer})$
            \State $\textit{reward} \gets \alpha (2\textit{r}-\textit{r}^*)-(1-\alpha)\times \tau(\mathbf{C}',\mathbf{C})$ 
            \State $Q(\textit{state,action}) \gets Q(\textit{state,action}) + \frac{1}{n}(\textit{reward}-Q(\textit{state,action}))$ \Comment{Update $Q$ table}
    \EndFor
    \State \textbf{return} $Q^*$
    \Comment{return trained $Q$ table}
  \end{algorithmic}
\end{algorithm}
\paragraph{Adaptive-$k$:} Identifying the number $k$ is crucial for context reduction as well as performance. Considering a fixed-$k$ might impact accuracy if $k<<n$. On the contrary, if $k\simeq n$, the token ratio will be higher, leading to higher costs. hence, to achieve minimal cost with maximum accuracy, the $k$ should be adaptive. To get a query-based adaptive context, we propose a lightweight $Q$-learning-based reinforcement learning algorithm that perceives an optimal policy for an agent operating in an environment. After training, we will have an optimal $Q$ table that takes the best action for a given state.
\[ 
\pi^* (state) = \underset{a}{\operatorname{argmax}}\;  Q^*(state,action) \] 
 
\paragraph{Details of RL:} Due to the dynamic environment of domains and queries, it is difficult to estimate the optimal context. In this situation where the environment is complex and dynamic, RL fits the best. After retrieving the context based on a query, \myTitle computes the state with context and query embedding. With the state, the RL agent finds a suitable action from the trained $Q^*$ table. Based on the action, the threshold for context reduction is computed and top-$k$ sentences are selected. Then the reduced context version is produced according to one of the sentence positioning variants. After then the response is retrieved from the reduced context and query and sent to the LLM.
We carefully define our state, action, and reward function as follows.


\paragraph{State:}  We combine query and context to define the \emph{state}. At offline profiling, we compute the embedding of the query $\mathbf{v}_{q}$ and the embedding of the context as $\mathbf{v}_{c}$. Then we subtract $\mathbf{v}_{q}$ from $\mathbf{v}_{c}$ that indicates the context-query pair. After building the vector with a number of training samples, we run the \textit{K-means} model to compute the centroids. These centroids are utilized as state vectors $\mathcal{S}$. So, at run time, a query context pair is concatenated using their embedding vectors and the closest centroid will be the state for them. 
\[
 \mathcal{S}\gets \textit{K-means}\left(\bigcup_{i,j}(\mathbf{v}_{c_i} - \mathbf{v}_{q_j})\right)
\]

\paragraph{Action:} As our goal is to make the top-k extraction adaptive. Given a set of thresholds from 0 to 0.4 (maximum top-$k$ will be 40\% of the total context) to select the number of sentences, we define each possible choice as an action of the proposed RL system. The outcome of each action choice is computed by the reward function and the $Q$ table is updated for the corresponding (state, action) pair. 

\paragraph{Reward:}
The reward for the RL model is higher if the context ratio is less and the accuracy is almost equal to the optimal accuracy using full context. We compute the ROUGE score (${r}$) to evaluate the answer using reduced context with the actual answer using full context. If the ROUGE score with expensive LLM is (${r}^*$) for a query, then the current (state, action) will be rewarded if ${r}-{r^*}\ge 0$, otherwise will be penalized. For the token ratio, the lower the better as the reduction of context as much as possible without compromising accuracy is rewarded. Thus, the reward function $\mathbf{R}$ is defined as follows.
\[
\mathbf{R} =  -(1-\alpha)\tau + \alpha(2{r}-{r^*})
\]

\paragraph{Training algorithm:} The off-policy $Q$ table training algorithm is shown in Algorithm~\ref{algo-ls}. In line $1-6$, each state is computed by the subtraction of query embedding from context embedding. A k-means model is trained to get the centroids and utilize the centroids as different states. In line $7-15$ for selecting each threshold as action, the corresponding reward is computed the $Q$ table is updated. Finally, the updated $Q$ table is deployed for reducing the context. The training of RL requires LLM to compute the reward. To reduce the training cost, we perform training on fewer samples. hence to observe the effect of each action in each state, we do a full exploration to update the $Q$ table.\par

\paragraph{\myTitle Inference}
The \myTitle inference algorithm is shown in Algorithm~\ref{algo-inference}. For each query, the corresponding context is retrieved, and the state is computed by the trained RL-agent [line 2]. Based on the state, the threshold $\theta$ is computed as an action to select the top-k sentences and the top-$k$ sentences with reduced less important sentences produce the reduced context, $\mathbf{C}'$ [line 3-4]. This reduced context is utilized to get answer with the LLM [line 5]. Finally, the answer is returned to the user.
\begin{algorithm}[H]
  \caption{\myTitle Inference}
  \label{algo-inference}
  \begin{algorithmic}[1]
    \Require{$\mathcal{D}_t$: Test documents}
    \Require{$\vect{q}_t$: A set of test queries}
    \Require{$Agent$: Trained RL-Agent}
    \For{each $q,\textit{context} \in \vect{q}_t,
    \mathcal{D}_t$}
        \State $\textit{state}\gets Agent.\text{get\_state}(Agent.\mathcal{S},\textit{context},q)$
        \State $action \gets Agent.\text{get\_action} (state)$ 
            \State $\mathbf{C}' \gets Agent.\text{perform\_action}(context,\textit{action})$ 
            \State $\textit{answer} \gets \textit{llm}(q,\mathbf{C}')$
        \State \textbf{return} $answer$
    \EndFor
  \end{algorithmic}
\end{algorithm}

\section{Experimental Settings}
\paragraph{Dataset:} In real-time scenarios, user documents are new to LLMs such as gpt-3.5-turbo model, and thus should not be able to answer a query without giving a context. To ensure this, we use recent arxiv papers and BBC news articles so that the LLMs are not trained on these. Following this, we use the Arxiv Dataset and BBC News Dataset where documents are published in March 2023~\cite{li2023unlocking}. We generate questions with answers for each document using QAGenerationChain from LangChain~\cite{langchain} based on gpt-3.5-turbo model.

    


\paragraph{Baseline Models:}\label{baseline} In our question-answering-based system, we mainly focus on reducing the context length while keeping the same performance in QA. So, we use a pay-per-use LLM ({\tt gpt-turbo-3.5}) model for all cases to answer a query based on the context. We evaluate our proposed context length reduction approach with the recent context reduction approaches as follows.
\begin{enumerate}
    \item Context (Original): We keep the context length intact and ask LLM to answer the query.
    
    \item CQSumDP~\cite{laskar2023cqsumdp}: We generate the following prompt similar to CQSumDP to generate the query-aware summary of a context.\\ 
    ``A document along with its query is given below. Write down the most reasonable summary relevant to its document-query pair.\\
    Document: \{CONTEXT\}\\
    Query: \{QUERY\}''
    
    \item Semantic Compression~\cite{gilbert2023semantic}: A query unaware prompt is stated to compress a context as follows.\\
        ``Please compress the following text into a latent representation that a different {\tt gpt-3.5-turbo} model can decompress into
    the original text. The compression model should purely
    minimize the number of characters in the compressed
    representation while maintaining the semantics of the
    original text. The resulting compressed text does not need
    to be decompressed into the original text but should
    capture the semantics of the original text. The compressed
    text should be able to be decompressed into a text that
    is semantically similar to the original text but does not
    need to be identical.\\
    Text to Compress: \{CONTEXT\}''
   
   \item SBert~\cite{miller2019leveraging}: A Bert-model-based extractive summarization approach. The context is reduced to several sentences. We keep the number of sentences 3 in our experiments.

    \item Selective Context (SC~\cite{li2023unlocking}): It utilizes entropy to filter out less informative content from context. In our experiments, we employ phrase-level content filtering with different reduction ratios. We use GPT-2 model to compute the self-information.

     \item Flan-T5-Base~\cite{chung2022scaling}: We use this 250M encoder-decoder model to summarize the context based on the same instruction as CQSumDP. 
\end{enumerate}

\paragraph{Implementation Details:} To implement the document ingestion, we use a cheap all-MiniLM-L6-v2 model~\cite{reimers2019sentence} as an embedding generator. The text chunks along with the embeddings are stored in ChromaDB~\cite{chromadb} vector database. We use the chunk size 500, chunk overlap is 0. We vary the number of chunks N = 2, 4, 8, 10 to compare how well the RL algorithm performs. We use this template for the QA: \emph{``Answer to the question based on the given context. Context: \{CONTEXT\}, 
Question: \{QUERY\}, if you do not find any answer in the context, simply return `No answer'"}. We use LLMChain from Langchain~\cite{langchain} to ask queries to the OpenAI model by prompting. \myTitle uses selective context method~\cite{li2023unlocking} to reduce less important sentences by 80\%.

\section{Results}
\paragraph{Arxiv Dataset:} We consider random 25 articles from the Arxiv dataset for testing and another 5 documents RL training distinct from the 25 documents. After retrieving the query, we use the RL agent to identify to top-k in the context and reduce the context using the RL agent. For the same query, and context setting, we evaluate our approach with the baseline approaches. The comparison of \myTitle with baseline models is shown in Table~\ref{tab:arxiv-results}. \myTitle achieves similar performance compared to the original context with no text reduction and outperforms existing open-source models with $37\%$ cost savings. CQSumDP~\cite{laskar2023cqsumdp} achieves better accuracy with a greater cost (adds 15.36\% more cost) by reducing the minimal context to get the right answer than the original context. Throughout the experiments, we observe the same effect and conclude that  the zero-shot performance of GPT-4 like LLMs is better with concise and relevant context to query (CQSumDP) than the context with a lot of irrelevant information. Although Semantic Compression~\cite{gilbert2023semantic} uses LLM to minimize context, due to the context-agnostic nature of the prompt, it performs even worse than \myTitle and adds additional $\sim72\%$ cost. 
\begin{table}[tbp]
\centering
\resizebox{\columnwidth}{!}{%
\begin{tabular}{@{}ccccccccc@{}}
\toprule
\rowcolor[HTML]{FFFFFF} 
\multicolumn{1}{c}{\cellcolor[HTML]{FFFFFF}{\color[HTML]{000000} \textbf{Text Reduction Method}}} &
  \multicolumn{1}{c}{\cellcolor[HTML]{FFFFFF}{\color[HTML]{000000} \textbf{\begin{tabular}[c]{@{}c@{}}Avg.\\ Total \\ tokens\end{tabular}}}} &
  \multicolumn{1}{c}{\cellcolor[HTML]{FFFFFF}{\color[HTML]{000000} \textbf{\begin{tabular}[c]{@{}c@{}}Avg.\\ Prompt \\ tokens\end{tabular}}}} &
  \multicolumn{1}{c}{\cellcolor[HTML]{FFFFFF}{\color[HTML]{000000} \textbf{\begin{tabular}[c]{@{}c@{}}Avg.\\ Summary \\ tokens\end{tabular}}}} &
  \multicolumn{1}{c}{\cellcolor[HTML]{FFFFFF}{\color[HTML]{000000} \textbf{\begin{tabular}[c]{@{}c@{}}Avg. \\ Completion\\ tokens\end{tabular}}}} &
  \multicolumn{1}{c}{\cellcolor[HTML]{FFFFFF}{\color[HTML]{000000} \textbf{ROUGE-1}}} &
  \multicolumn{1}{c}{\cellcolor[HTML]{FFFFFF}{\color[HTML]{000000} \textbf{ROUGE-2}}} &
  \multicolumn{1}{c}{\cellcolor[HTML]{FFFFFF}{\color[HTML]{000000} \textbf{ROUGE-L}}} &
  \multicolumn{1}{c}{\cellcolor[HTML]{FFFFFF}{\color[HTML]{000000} \textbf{\begin{tabular}[c]{@{}c@{}}Cost Savings \\ (\%)\end{tabular}}}} \\ \midrule
\rowcolor[HTML]{D4FAF8} 
{\color[HTML]{000000} Context (Original)} &
  {\color[HTML]{000000} 547} &
  {\color[HTML]{000000} 521} &
  {\color[HTML]{000000} 0} &
  {\color[HTML]{000000} 26} &
  {\color[HTML]{000000} \textbf{0.3985}} &
  {\color[HTML]{000000} \textbf{0.2868}} &
  {\color[HTML]{000000} \textbf{0.3714}} &
  {\color[HTML]{000000} 0.00} \\
{\color[HTML]{000000} CQSumDP} &
  {\color[HTML]{000000} 631} &
  {\color[HTML]{000000} 89} &
  {\color[HTML]{000000} 517} &
  {\color[HTML]{000000} 25} &
  {\color[HTML]{000000} \textbf{0.4424}} &
  {\color[HTML]{000000} \textbf{0.3061}} &
  {\color[HTML]{000000} \textbf{0.4213}} &
  {\color[HTML]{FE0000} \textbf{-15.36}} \\
{\color[HTML]{000000} Semantic Compression} &
  {\color[HTML]{000000} 939} &
  {\color[HTML]{000000} 319} &
  {\color[HTML]{000000} 597} &
  {\color[HTML]{000000} 23} &
  {\color[HTML]{000000} 0.3331} &
  {\color[HTML]{000000} 0.2221} &
  {\color[HTML]{000000} 0.3132} &
  {\color[HTML]{000000} -71.66} \\
{\color[HTML]{000000} T5-base} &
  {\color[HTML]{000000} 79} &
  {\color[HTML]{000000} 71} &
  {\color[HTML]{000000} 0} &
  {\color[HTML]{000000} 8} &
  {\color[HTML]{000000} 0.1614} &
  {\color[HTML]{000000} 0.1101} &
  {\color[HTML]{000000} 0.1486} &
  {\color[HTML]{000000} 85.56} \\
{\color[HTML]{000000} SBert} &
  {\color[HTML]{000000} 205} &
  {\color[HTML]{000000} 188} &
  {\color[HTML]{000000} 0} &
  {\color[HTML]{000000} 17} &
  {\color[HTML]{000000} 0.2563} &
  {\color[HTML]{000000} 0.1701} &
  {\color[HTML]{000000} 0.2469} &
  {\color[HTML]{000000} 62.52} \\
{\color[HTML]{000000} SC (reduction = 0.50)} &
  {\color[HTML]{000000} 334} &
  {\color[HTML]{000000} 316} &
  {\color[HTML]{000000} 0} &
  {\color[HTML]{000000} 18} &
  {\color[HTML]{000000} 0.2945} &
  {\color[HTML]{000000} 0.2014} &
  {\color[HTML]{000000} 0.2755} &
  {\color[HTML]{000000} 38.94} \\\midrule
{\color[HTML]{000000} LeanContext (Fixed k =0.1)} &
  {\color[HTML]{000000} 210} &
  {\color[HTML]{000000} 196} &
  {\color[HTML]{000000} 0} &
  {\color[HTML]{000000} 14} &
  {\color[HTML]{000000} 0.2305} &
  {\color[HTML]{000000} 0.1623} &
  {\color[HTML]{000000} 0.2173} &
  {\color[HTML]{000000} 61.62} \\
{\color[HTML]{213AF8} \textbf{LeanContext (Adaptive k {[}RL{]})}} &
  {\color[HTML]{000000} \textbf{343}} &
  {\color[HTML]{000000} \textbf{321}} &
  {\color[HTML]{000000} \textbf{0}} &
  {\color[HTML]{000000} \textbf{22}} &
  {\color[HTML]{3531FF} \textbf{0.3844}} &
  {\color[HTML]{3531FF} \textbf{0.2684}} &
  {\color[HTML]{3531FF} \textbf{0.3577}} &
  {\color[HTML]{3531FF} \textbf{37.29}}
  \\
  \bottomrule
\end{tabular}%
}
\caption{Comparison on Random 100 samples from Arxiv Dataset. Number of chunks, $N=4$}
\label{tab:arxiv-results}
\end{table}


\paragraph{BBCNews Dataset:} We consider random 100 news articles from the BBCNews dataset. We keep 80 articles for testing and the rest 20 articles to train the RL agent. For query-based context retrieval settings, we evaluate our approach with the baseline approaches.

Due to the cost of the OpenAI model and the current usage limit, we follow recent literature~\cite{yang2023exploring}, we generate queries using the QA generation method, and consider random 100 query samples for evaluation. The evaluation result is shown in Table~\ref{tab:bbcnews-eval}.

\begin{table}[tbp]
\centering
\resizebox{\columnwidth}{!}{%
\begin{tabular}{@{}ccccccccc@{}}
\toprule
{\color[HTML]{000000} \textbf{Text Reduction Method}} &
  {\color[HTML]{000000} \textbf{\begin{tabular}[c]{@{}c@{}}Avg.\\ Total\\ tokens\end{tabular}}} &
  {\color[HTML]{000000} \textbf{\begin{tabular}[c]{@{}c@{}}Avg.\\ Prompt\\ tokens\end{tabular}}} &
  {\color[HTML]{000000} \textbf{\begin{tabular}[c]{@{}c@{}}Avg.\\ Summary\\tokens\end{tabular}}} &
  {\color[HTML]{000000} \textbf{\begin{tabular}[c]{@{}c@{}}Avg. \\ Completion\\tokens\end{tabular}}} &
  {\color[HTML]{000000} \textbf{ROUGE-1}} &
  {\color[HTML]{000000} \textbf{ROUGE-2}} &
  {\color[HTML]{000000} \textbf{ROUGE-L}} &
  {\color[HTML]{000000} \textbf{\begin{tabular}[c]{@{}c@{}}Cost Savings\\(\%)\end{tabular}}} \\ \midrule
  \rowcolor[HTML]{D4FAF8} 
{\color[HTML]{000000} Context (Original)} &
  {\color[HTML]{000000} 761} &
  {\color[HTML]{000000} 724} &
  {\color[HTML]{000000} 0} &
  {\color[HTML]{000000} 37} &
  {\color[HTML]{000000} \textbf{0.5498}} &
  {\color[HTML]{000000} \textbf{0.4172}} &
  {\color[HTML]{000000} \textbf{0.5337}} &
  0.00 \\
{\color[HTML]{000000} CQSumDP} &
  {\color[HTML]{000000} 842} &
  {\color[HTML]{000000} 75} &
  {\color[HTML]{000000} 738} &
  {\color[HTML]{000000} 29} &
  {\color[HTML]{000000} \textbf{0.5801}} &
  {\color[HTML]{000000} \textbf{0.4405}} &
  {\color[HTML]{000000} \textbf{0.5637}} &
  {\color[HTML]{FE0000} \textbf{-10.64}} \\
{\color[HTML]{000000} Semantic Compression} &
  {\color[HTML]{000000} 1078} &
  {\color[HTML]{000000} 228} &
  {\color[HTML]{000000} 820} &
  {\color[HTML]{000000} 30} &
  {\color[HTML]{000000} 0.4729} &
  {\color[HTML]{000000} 0.3204} &
  {\color[HTML]{000000} 0.4517} &
  -41.66 \\
{\color[HTML]{000000} T5-base} &
  {\color[HTML]{000000} 74} &
  {\color[HTML]{000000} 51} &
  {\color[HTML]{000000} 0} &
  {\color[HTML]{000000} 23} &
  {\color[HTML]{000000} 0.3993} &
  {\color[HTML]{000000} 0.2631} &
  {\color[HTML]{000000} 0.3752} &
  90.28 \\
SBert &
  174 &
  147 &
  0 &
  27 &
  0.4261 &
  0.2917 &
  0.4082 &
  77.14 \\
SC (reduction = 0.50) &
  461 &
  429 &
  0 &
  32 &
  0.4740 &
  0.3308 &
  0.4521 &
  39.42 \\\midrule
{\color[HTML]{000000}LeanContext (Fixed k = 0.1)}&
  {\color[HTML]{000000}\textbf{278}} &
  {\color[HTML]{000000}\textbf{250}} &
  {\color[HTML]{000000}\textbf{0}} &
  {\color[HTML]{000000}\textbf{28}} &
  {\color[HTML]{000000}\textbf{0.5017}} &
  {\color[HTML]{000000}\textbf{0.3740}} &
  {\color[HTML]{000000}\textbf{0.4872}} &
  {\color[HTML]{000000}\textbf{63.47} }\\
{\color[HTML]{213AF8}\textbf{LeanContext (Adaptive-k {[}RL{]})}} &
  {\color[HTML]{000000}\textbf{245}} &
  {\color[HTML]{000000}\textbf{218}} &
  {\color[HTML]{000000}\textbf{0}} &
  {\color[HTML]{000000}\textbf{27}} &
  
  {\color[HTML]{3531FF}\textbf{0.5233}} &
  {\color[HTML]{3531FF}\textbf{0.3943}} &
  {\color[HTML]{3531FF}\textbf{0.5093}} &
  {\color[HTML]{3531FF}\textbf{67.81}} \\ \bottomrule
\end{tabular}%
}
\caption{Comparison on Random 100 samples from BBCNews Dataset. Number of chunks, $N=8$}
\label{tab:bbcnews-eval}
\end{table}

We observe that generating a query-aware summary from document-query pair using OpenAI LLM~\cite{laskar2023cqsumdp} performs better i.e. ROUGE-1 0.5801 than query-aware open-source baseline methods such as T5-base i.e. 0.3993 but with a high cost. However, it also contributes more cost than the original context ($10.64\%$ more). We also observe that the query-unaware LLM using a different hard prompt~\cite{gilbert2023semantic} even performs worse i.e. ROUGE-1 score 0.4729 with a 41.66\% cost overhead. Additionally, adding top-$k$(0.1) uplifts the ROUGE-1 score of T5-base by $\sim12.35\%$. With adaptive top-$k$ (Top-$k$(RL)) and adaptive top-$k$ (Top-$k$(RL)) with sentence order, it performs even better by reducing the cost from $65\%\sim74\%$. We observe the same scenario for other models too. Further investigation with different number of chunks for the same test queries are discussed in the Ablation Study section.



\paragraph{Top-k is all you need:}
We observe an interesting phenomenon when adding our \myTitle with existing open-source models. As these open-source models are not trained on new domain data, we observe that combining $10\%$ top-$k$ sentences with the open-source models boosts the QA system's performance by $5.41\%\sim17.11\%$ for the Arxiv dataset and $8.11\%\sim12.35\%$ for the BBCNews dataset as shown in Table~\ref{tab:top-kadvg}. In addition, using top-$k$ on the fly with a generic summarizer adds an extra advantage to the overall system. Our \myTitle with top-$k$ sentences using RL and reduced version of other sentences in between them make the performance even better than the fixed ($10\%$) top-$k$ by $13.22\%\sim24.59\%$ for the Arxiv dataset and $10.94\%\sim15.39\%$ for the BBCNews dataset (Table~\ref{tab:top-kadvg}). \myTitle with T5-base model outperforms all the existing approaches including no reduction method.

\begin{table}[!tbp]
\centering
\resizebox{\columnwidth}{!}{%
\begin{tabular}{@{}lccccccccc@{}}
\toprule
\rowcolor[HTML]{FFFFFF} 
{\color[HTML]{000000} \textbf{Dataset}} &
  {\color[HTML]{000000} \textbf{Text Reduction Method}} &
  {\color[HTML]{000000} \textbf{\begin{tabular}[c]{@{}c@{}}Avg.\\ Total \\ tokens\end{tabular}}} &
  {\color[HTML]{000000} \textbf{\begin{tabular}[c]{@{}c@{}}Avg.\\ Prompt \\ tokens\end{tabular}}} &
  {\color[HTML]{000000} \textbf{\begin{tabular}[c]{@{}c@{}}Avg.\\ Summary \\ tokens\end{tabular}}} &
  {\color[HTML]{000000} \textbf{\begin{tabular}[c]{@{}c@{}}Avg. \\ Completion\\ tokens\end{tabular}}} &
  {\color[HTML]{000000} \textbf{ROUGE-1}} &
  {\color[HTML]{000000} \textbf{ROUGE-2}} &
  {\color[HTML]{000000} \textbf{ROUGE-L}} &
  {\color[HTML]{000000} \textbf{\begin{tabular}[c]{@{}c@{}}Cost Savings \\ (\%)\end{tabular}}} \\ \midrule
\rowcolor[HTML]{D4FAF8} 
\cellcolor[HTML]{FFFFFF} &
  {\color[HTML]{000000} \textbf{Context (Original)}} &
  {\color[HTML]{000000} \textbf{547}} &
  {\color[HTML]{000000} \textbf{521}} &
  {\color[HTML]{000000} \textbf{0}} &
  {\color[HTML]{000000} \textbf{26}} &
  {\color[HTML]{000000} \textbf{0.3985}} &
  {\color[HTML]{000000} \textbf{0.2868}} &
  {\color[HTML]{000000} \textbf{0.3714}} &
  {\color[HTML]{000000} \textbf{0.00}} \\
\rowcolor[HTML]{FFFFFF} 
 &
  \textbf{T5} &
  {\color[HTML]{000000} 79} &
  {\color[HTML]{000000} 71} &
  {\color[HTML]{000000} 0} &
  {\color[HTML]{000000} 8} &
  {\color[HTML]{000000} 0.1614} &
  {\color[HTML]{000000} 0.1101} &
  {\color[HTML]{000000} 0.1486} &
  {\color[HTML]{000000} 85.56} \\
\rowcolor[HTML]{FFFFFF} 
 &
  \textbf{T5 + LeanContext (Only Top-k=0.1)} &
  {\color[HTML]{000000} 131} &
  {\color[HTML]{000000} 113} &
  {\color[HTML]{000000} 0} &
  {\color[HTML]{000000} 18} &
  {\color[HTML]{000000} 0.3325} &
  {\color[HTML]{000000} 0.2390} &
  {\color[HTML]{000000} 0.3146} &
  {\color[HTML]{000000} 76.05} \\ 
\rowcolor[HTML]{FFFFFF} 
 &
  \textbf{T5 + LeanContext (Only Top-k=RL)} &
  {\color[HTML]{000000} 284} &
  {\color[HTML]{000000} 263} &
  {\color[HTML]{000000} 0} &
  {\color[HTML]{000000} 21} &
  {\color[HTML]{000000} 0.3942} &
  {\color[HTML]{000000} 0.2847} &
  {\color[HTML]{000000} 0.3742} &
  {\color[HTML]{000000} 48.08} \\
\rowcolor[HTML]{FFFFFF} 
\textbf{Arxiv} &
  {\color[HTML]{213AF8} \textbf{T5 + LeanContext}} &
  \textbf{357} &
  \textbf{335} &
  \textbf{0} &
  \textbf{22} &
  {\color[HTML]{FF0000} \textbf{0.4073}} &
  {\color[HTML]{FF0000} \textbf{0.2863}} &
  {\color[HTML]{FF0000} \textbf{0.3809}} &
  \textbf{34.73} \\\cmidrule(l){2-10}
\rowcolor[HTML]{FFFFFF} 
 &
  \textbf{SBert} &
  {\color[HTML]{000000} 205} &
  {\color[HTML]{000000} 188} &
  {\color[HTML]{000000} 0} &
  {\color[HTML]{000000} 17} &
  {\color[HTML]{000000} 0.2563} &
  {\color[HTML]{000000} 0.1701} &
  {\color[HTML]{000000} 0.2469} &
  {\color[HTML]{000000} 62.52} \\
\rowcolor[HTML]{FFFFFF} 
 &
  \textbf{SBert + LeanContext (Only Top-k=0.1)} &
  {\color[HTML]{000000} 250} &
  {\color[HTML]{000000} 230} &
  {\color[HTML]{000000} 0} &
  {\color[HTML]{000000} 20} &
  {\color[HTML]{000000} 0.3104} &
  {\color[HTML]{000000} 0.2181} &
  {\color[HTML]{000000} 0.2949} &
  {\color[HTML]{000000} 54.30} \\
\rowcolor[HTML]{FFFFFF} 
 &
  \textbf{SBert + LeanContext (Only Top-k=RL)} &
  {\color[HTML]{000000} 405} &
  {\color[HTML]{000000} 380} &
  {\color[HTML]{000000} 0} &
  {\color[HTML]{000000} 25} &
  {\color[HTML]{000000} 0.3676} &
  {\color[HTML]{000000} 0.2594} &
  {\color[HTML]{000000} 0.3464} &
  {\color[HTML]{000000} 25.96} \\
\rowcolor[HTML]{FFFFFF} 
 &
  {\color[HTML]{213AF8} \textbf{SBert + LeanContext}} &
  \textbf{478} &
  \textbf{452} &
  \textbf{0} &
  \textbf{26} &
  {\color[HTML]{FF0000} \textbf{0.3885}} &
  {\color[HTML]{FF0000} \textbf{0.2720}} &
  {\color[HTML]{FF0000} \textbf{0.3596}} &
  \textbf{12.61} \\ \midrule
\rowcolor[HTML]{D4FAF8} 
\cellcolor[HTML]{FFFFFF} &
  {\color[HTML]{000000} \textbf{Context (Original)}} &
  {\color[HTML]{000000} \textbf{761}} &
  {\color[HTML]{000000} \textbf{724}} &
  {\color[HTML]{000000} \textbf{0}} &
  {\color[HTML]{000000} \textbf{37}} &
  {\color[HTML]{000000} \textbf{0.5498}} &
  {\color[HTML]{000000} \textbf{0.4172}} &
  {\color[HTML]{000000} \textbf{0.5337}} &
  {\color[HTML]{000000} \textbf{0}} \\
\rowcolor[HTML]{FFFFFF} 
 &
  \textbf{T5} &
  {\color[HTML]{000000} 74} &
  {\color[HTML]{000000} 51} &
  {\color[HTML]{000000} 0} &
  {\color[HTML]{000000} 23} &
  {\color[HTML]{000000} 0.3993} &
  {\color[HTML]{000000} 0.2631} &
  {\color[HTML]{000000} 0.3752} &
  {\color[HTML]{000000} 90.28} \\
\rowcolor[HTML]{FFFFFF} 
 &
  \textbf{T5 + LeanContext (Only Top-k=0.1)} &
  {\color[HTML]{000000} 142} &
  {\color[HTML]{000000} 116} &
  {\color[HTML]{000000} 0} &
  {\color[HTML]{000000} 26} &
  {\color[HTML]{000000} 0.5228} &
  {\color[HTML]{000000} 0.3914} &
  {\color[HTML]{000000} 0.5065} &
  {\color[HTML]{000000} 81.34} \\
\rowcolor[HTML]{FFFFFF} 
 &
  \textbf{T5 + LeanContext (Only Top-k=RL) } &
  {\color[HTML]{000000} 192} &
  {\color[HTML]{000000} 165} &
  {\color[HTML]{000000} 0} &
  {\color[HTML]{000000} 27} &
  {\color[HTML]{000000} 0.5368} &
  {\color[HTML]{000000} 0.4072} &
  {\color[HTML]{000000} 0.5187} &
  {\color[HTML]{000000} 74.77} \\
\rowcolor[HTML]{FFFFFF} 
\textbf{BBCNews} &
  {\color[HTML]{213AF8} \textbf{T5 + LeanContext}} &
  \textbf{259} &
  \textbf{232} &
  \textbf{0} &
  \textbf{27} &
  {\color[HTML]{FF0000} \textbf{0.5532}} &
  {\color[HTML]{FF0000} \textbf{0.4298}} &
  {\color[HTML]{FF0000} \textbf{0.5374}} &
  \textbf{65.97} \\\cmidrule(l){2-10} 
\rowcolor[HTML]{FFFFFF} 
 &
  \textbf{SBert} &
  {\color[HTML]{000000} 174} &
  {\color[HTML]{000000} 147} &
  {\color[HTML]{000000} 0} &
  {\color[HTML]{000000} 27} &
  {\color[HTML]{000000} 0.4261} &
  {\color[HTML]{000000} 0.2917} &
  {\color[HTML]{000000} 0.4082} &
  {\color[HTML]{000000} 77.14} \\
\rowcolor[HTML]{FFFFFF} 
 &
  \textbf{SBert + LeanContext (Only Top-k=0.1)} &
  {\color[HTML]{000000} 241} &
  {\color[HTML]{000000} 212} &
  {\color[HTML]{000000} 0} &
  {\color[HTML]{000000} 29} &
  {\color[HTML]{000000} 0.5072} &
  {\color[HTML]{000000} 0.3731} &
  {\color[HTML]{000000} 0.4914} &
  {\color[HTML]{000000} 68.33} \\
\rowcolor[HTML]{FFFFFF} 
 &
  \textbf{SBert + LeanContext (Only Top-k=RL)} &
  {\color[HTML]{000000} 289} &
  {\color[HTML]{000000} 260} &
  {\color[HTML]{000000} 0} &
  {\color[HTML]{000000} 29} &
  {\color[HTML]{000000} 0.5200} &
  {\color[HTML]{000000} 0.3827} &
  {\color[HTML]{000000} 0.5039} &
  {\color[HTML]{000000} 62.02} \\
\rowcolor[HTML]{FFFFFF} 
 &
  {\color[HTML]{213AF8} \textbf{SBert + LeanContext}} &
  \textbf{355} &
  \textbf{327} &
  \textbf{0} &
  \textbf{28} &
  {\color[HTML]{FF0000} \textbf{0.5355}} &
  {\color[HTML]{FF0000} \textbf{0.4007}} &
  {\color[HTML]{FF0000} \textbf{0.5235}} &
  \textbf{53.36} \\ \bottomrule
\end{tabular}%
}
\caption{Effect of cascading LeanContext with open source summarizers.}
\label{tab:top-kadvg}
\end{table}

\begin{table}[!h]
\centering
 

\resizebox{\linewidth}{!}{%
\begin{tabular}{lcccccccc}
\toprule
\textbf{Text Reduction Method} &
  \textbf{\begin{tabular}[c]{@{}c@{}}Avg. \\ Total \\ tokens\end{tabular}} &
  \textbf{\begin{tabular}[c]{@{}c@{}}Avg. \\ Prompt \\ tokens\end{tabular}} &
  \textbf{\begin{tabular}[c]{@{}c@{}}Avg. \\ Summary \\ tokens\end{tabular}} &
  \textbf{\begin{tabular}[c]{@{}c@{}}Avg. \\ Completion \\ tokens\end{tabular}} &
  \textbf{ROUGE-1} &
  \textbf{ROUGE-2} &
  \textbf{ROUGE-L} &
  \textbf{\begin{tabular}[c]{@{}c@{}}Cost Savings\\ (\%)\end{tabular}} \\
  \rowcolor[HTML]{F8FF00}
  \toprule
\textbf{Number of chunks, N=2}& &  &&& & & &\\

Context (Original) & 
  243 &
  211 &
  0 &
  32 &
  {\color[HTML]{32CB00} \textbf{0.5370}} &
  {\color[HTML]{32CB00} \textbf{0.3987}} &
  {\color[HTML]{32CB00} \textbf{0.5190}} &
  0.00 \\
CQSumDP &
  322 &
  70 &
  225 &
  27 &
  {\color[HTML]{32CB00} \textbf{0.5303}} &
  {\color[HTML]{32CB00} \textbf{0.3897}} &
  {\color[HTML]{32CB00} \textbf{0.5105}} &
  -32.51 \\
Semantic Compression &
  448 &
  113 &
  307 &
  28 &
  0.4786 &
  0.3169 &
  0.4519 &
  -84.36 \\
T5-base &
  76 &
  50 &
  0 &
  26 &
  {\color[HTML]{343434} 0.4010} &
  {\color[HTML]{343434} 0.2640} &
  {\color[HTML]{343434} 0.3781} &
  68.72 \\
SBert &
  157 &
  128 &
  0 &
  29 &
  {\color[HTML]{343434} 0.4825} &
  {\color[HTML]{343434} 0.3395} &
  {\color[HTML]{343434} 0.4604} &
  35.39 \\
SC (reduction = 0.50) &
  164 &
  137 &
  0 &
  27 &
  {\color[HTML]{343434} 0.4614} &
  {\color[HTML]{343434} 0.3113} &
  {\color[HTML]{343434} 0.4403} &
  {\color[HTML]{343434} 32.51} \\\midrule
{\color[HTML]{213AF8}\textbf{LeanContext}} &
  117 &
  92 &
  0 &
  25 &
  {\color[HTML]{FE0000} \textbf{0.4556}} &
  {\color[HTML]{FE0000} \textbf{0.3265}} &
  {\color[HTML]{FE0000} \textbf{0.4373}} &
   \textbf{51.85}\\
\bottomrule

\bottomrule
 
\rowcolor[HTML]{F8FF00}
\textbf{Number of chunks, N=4}& &  &&& & & &\\

Context (Original) &
  417 &
  384 &
  0 &
  33 &
  {\color[HTML]{32CB00} \textbf{0.5489}} &
  {\color[HTML]{32CB00} \textbf{0.4183}} &
  {\color[HTML]{32CB00} \textbf{0.5327}} &
  0.00 \\
CQSumDP &
  496 &
  71 &
  398 &
  27 &
  {\color[HTML]{32CB00} \textbf{0.5569}} &
  {\color[HTML]{32CB00} \textbf{0.4156}} &
  {\color[HTML]{32CB00} \textbf{0.5389}} &
  -18.94 \\
Semantic Compression &
  665 &
  159 &
  479 &
  27 &
  0.4916 &
  0.3328 &
  0.4684 &
  -59.47 \\
T5-base &
  74 &
  50 &
  0 &
  24 &
  {\color[HTML]{343434} 0.3921} &
  {\color[HTML]{343434} 0.2555} &
  {\color[HTML]{343434} 0.3678} &
  82.25 \\
SBert &
  163 &
  136 &
  0 &
  27 &
  {\color[HTML]{343434} 0.4626} &
  {\color[HTML]{343434} 0.3287} &
  {\color[HTML]{343434} 0.4452} &
  \textbf{60.91} \\
SC (reduction = 0.50) &
  263 &
  235 &
  0 &
  28 &
  {\color[HTML]{343434} 0.4726} &
  {\color[HTML]{343434} 0.3250} &
  {\color[HTML]{343434} 0.4507} &
  {\color[HTML]{343434} 36.93} \\\midrule
{\color[HTML]{213AF8}\textbf{LeanContext}} &
  153 &
  128 &
  0 &
  25 &
  {\color[HTML]{FE0000} \textbf{0.4856}} &
  {\color[HTML]{FE0000} \textbf{0.3605}} &
  {\color[HTML]{FE0000} \textbf{0.4700}} &
  \textbf{63.31}\\
  \bottomrule



\bottomrule
\rowcolor[HTML]{F8FF00}
 \textbf{Number of chunks, N=10}& &  &&& & & &\\
  
Context (Original) &
  930 &
  892 &
  0 &
  38 &
  {\color[HTML]{32CB00} \textbf{0.5423}} &
  {\color[HTML]{32CB00} \textbf{0.4072}} &
  {\color[HTML]{32CB00} \textbf{0.5246}} &
  0.00 \\
CQSumDP &
  1008 &
  74 &
  906 &
  28 &
  {\color[HTML]{32CB00} \textbf{0.5849}} &
  {\color[HTML]{32CB00} \textbf{0.4440}} &
  {\color[HTML]{32CB00} \textbf{0.5644}} &
  -8.39 \\
Semantic Compression &
  1274 &
  258 &
  987 &
  29 &
  0.4862 &
  0.3343 &
  0.4642 &
  -36.99 \\

T5-base &
  74 &
  51 &
  0 &
  23 &
  {\color[HTML]{343434} 0.3903} &
  {\color[HTML]{343434} 0.2584} &
  {\color[HTML]{343434} 0.3688} &
  92.04 \\

SBert &
  181 &
  153 &
  0 &
  28 &
  {\color[HTML]{343434} 0.4164} &
  {\color[HTML]{343434} 0.2807} &
  {\color[HTML]{343434} 0.3967} &
  \textbf{80.54} \\

SC (reduction = 0.50) &
  559 &
  527 &
  0 &
  32 &
  {\color[HTML]{343434} 0.4825} &
  {\color[HTML]{343434} 0.3312} &
  {\color[HTML]{343434} 0.4591} &
  {\color[HTML]{343434} 39.89} \\
\midrule
{\color[HTML]{213AF8}\textbf{LeanContext}} &
  279 &
  252 &
  0 &
  27 &
  {\color[HTML]{FE0000} \textbf{0.5318}} &
  {\color[HTML]{FE0000} \textbf{0.4069}} &
  {\color[HTML]{FE0000} \textbf{0.5205}} &
   \textbf{70.00}\\
  \bottomrule
\end{tabular}%
}
\caption{Evaluation of RL model on a different number of chunks on a random 100 samples from BBCNews Dataset. Our RL agent trained with (N=8) shows promising results while applying on different chunk numbers.}
\label{tab:diff-N}
\end{table}

\section{Ablation Study}
\begin{figure}[!t]
    \centering
    \includegraphics[width=0.8\linewidth]{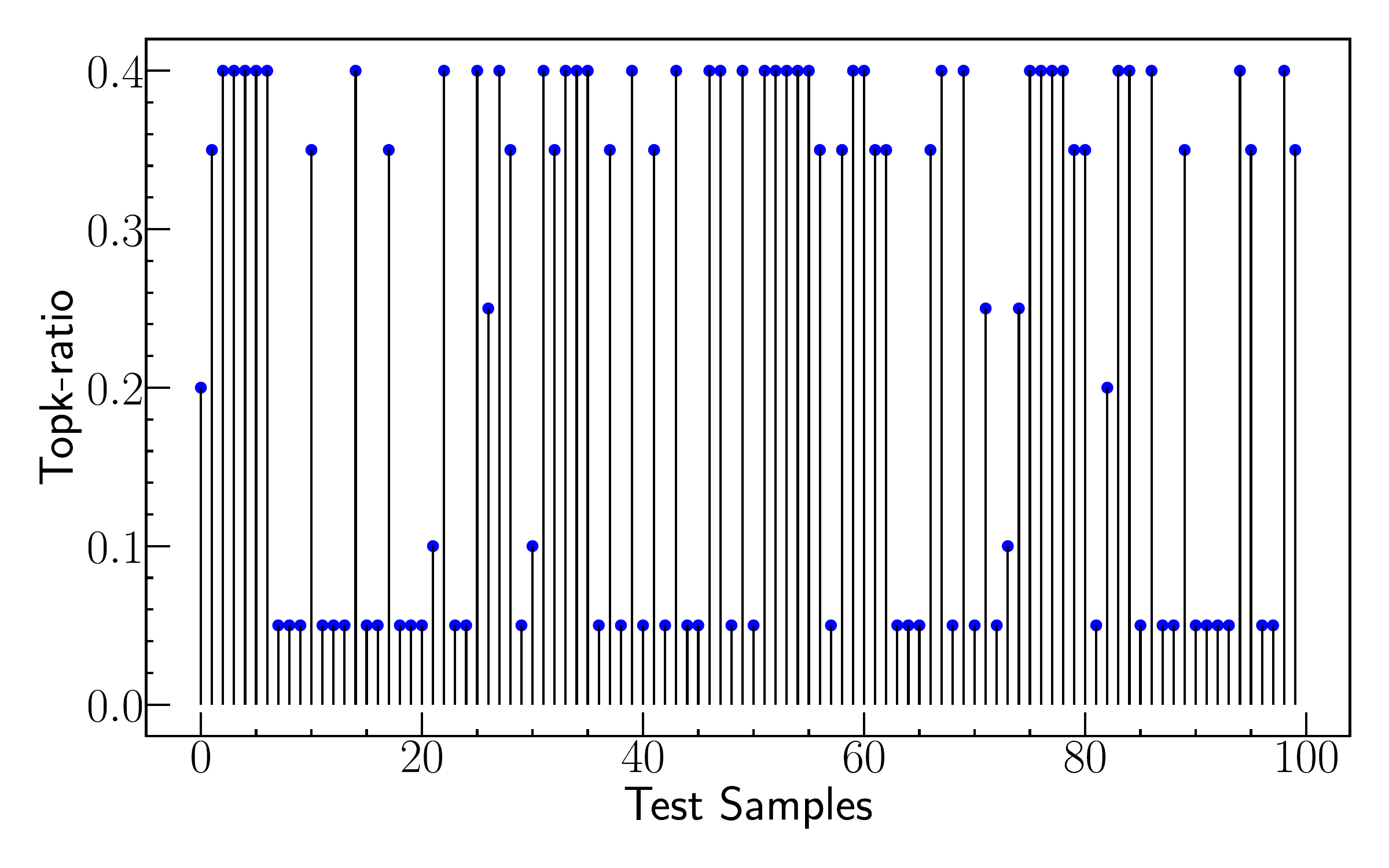}
    \caption{Adaptive-$k$ ratio selected by the BBCNews RL agent based on queries.}
    \label{fig:adaptive-k}
\end{figure}

 We investigate whether the RL model has a performance dependency on the number of chunks. So, we train the RL model using $N=8$ and run the same model on $N=2, 4,$ and $10$. The results are shown in Table~\ref{tab:diff-N}. We observe that although the $N$ changes, the RL model still outperforms other less expensive baseline methods. Another interesting observation lies in making N as larger as possible for the successful retrieval of context using \myTitle whereas previously without \myTitle $N$ should be kept smaller to reduce the cost of LLM usage.
In Figure~\ref{fig:adaptive-k}, we show how the action (top-$k$ ratio) is taken by the RL agent given the context and query for each query sample out of 100 samples. Considering an adaptive top-k ratio chosen by our RL agent varies over query samples to achieve the adaptive reduction of context. 

We also empirically evaluate state function with different associations between context ($\vect{v}_c$) and query embedding ($\vect{v}_q$) i.e. cosine similarity $\left(\text{state}= \frac{\vect{v}_c\cdot \vect{v}_q}{||\vect{v}_c|| ||\vect{v}_q||}\right)$, concatenation ($\vect{v}_c \oplus \vect{v}_q)$), and subtraction ($\vect{v}_c - \vect{v}_q)$). Among them, subtraction performs best (Table~\ref{tab:states}). In this experiment, we only select the top-k sentences using RL to evaluate the impact of state definition on performance.

\begin{table}[!t]
\centering
\resizebox{\linewidth}{!}{%
\begin{tabular}{@{}lcccccc@{}}
\toprule
\textbf{Compression Method} &
  \textbf{\begin{tabular}[c]{@{}c@{}}Avg. \\ Total \\ tokens\end{tabular}} &
  \textbf{\begin{tabular}[c]{@{}c@{}}Avg. \\ Prompt \\ tokens\end{tabular}} &
  \textbf{\begin{tabular}[c]{@{}c@{}}Avg. \\ Completion \\ tokens\end{tabular}} &
  \textbf{ROUGE-1} &
  \textbf{ROUGE-2} &
  \textbf{ROUGE-L} \\ \midrule
None                     & 538 & 514 & 24 & {\color[HTML]{32CB00} \textbf{0.3945}} & {\color[HTML]{32CB00} \textbf{0.2904}} & {\color[HTML]{32CB00} \textbf{0.3764}} \\
LeanContext $\left(\text{state}= \frac{\vect{v}_c\cdot \vect{v}_q}{||\vect{v}_c|| ||\vect{v}_q||}\right)$   & 296 & 275 & 21 & {\color[HTML]{000000} 0.3443}          & {\color[HTML]{000000} 0.2369}          & {\color[HTML]{000000} 0.3259}          \\
LeanContext ($\text{state}=\vect{v}_c-\vect{v}_q)$ & 331 & 308 & 23 & {\color[HTML]{FE0000} \textbf{0.3553}} & {\color[HTML]{FE0000} \textbf{0.2535}} & {\color[HTML]{FE0000} \textbf{0.3388}} \\
LeanContext ($\text{state}=\vect{v}_c \oplus \vect{v}_q)$   & 284 & 265 & 19 & {\color[HTML]{000000} 0.3400} & {\color[HTML]{000000} 0.2370} & {\color[HTML]{000000} 0.3223} \\ \bottomrule
\end{tabular}%
}
\caption{Comparison of RL state functions on Arxiv dataset}
\label{tab:states}
\end{table}

\section{Conclusion and Future Work}
In this paper, we propose \myTitle, a cost-efficient query-aware context reduction system to reduce the cost associated with LLM API usage. Despite the reduction of prompt tokens, \myTitle achieves similar or better performance compared to the no reduction of the original context. The advantage of top-$k$ is that it can be plugged in with any existing summarization method of a domain-based QA system to boost the overall performance according to our experimental results. Here, we only focus on text-based context as a domain. We will explore other domains in our future work.

\bibliography{aaai24}

\end{document}